\documentclass{article}
\usepackage{arxiv}
\usepackage{array}
\pdfoutput=1
\usepackage{float}
\usepackage{listings}
\usepackage{pifont}
\usepackage{amssymb}
\usepackage{amsmath}
\usepackage{booktabs}
\usepackage{enumitem}
\usepackage{array}
\usepackage{float}
\usepackage{listings}
\usepackage{pifont}
\usepackage{amssymb}
\usepackage{amsmath}
\usepackage{multirow}
\usepackage{caption}
\usepackage[pdftex]{graphicx}
\usepackage{natbib}
\usepackage{booktabs}
\usepackage{subcaption}
\usepackage{tabularx}
\newcolumntype{Y}{>{\raggedright\arraybackslash}X}
\usepackage{multirow}
\usepackage[pdftex]{graphicx}
\usepackage{natbib}
\usepackage{todonotes}

\usetikzlibrary{shapes}

\usepackage[utf8]{inputenc} %
\usepackage[T1]{fontenc}    %
\usepackage{hyperref}       %
\usepackage{url}            %
\usepackage{booktabs}       %
\usepackage{amsfonts}       %
\usepackage{nicefrac}       %
\usepackage{microtype}      %
\usepackage{lipsum}
\usepackage{graphicx}
\graphicspath{ {./images/} }

\title{Clique Constraints for Exact Causal Structure Learning}

\author{
 Xiaoyu He \\
 AI Center \\
    Czech Technical University in Prague \\
    Karlovo namesti 13, Prague 12135\\
  \texttt{hexiaoyu@fel.cvut.cz} \\
   \And
 Pavel Ryt{\'i}{\v{r}} \\
 AI Center \\
    Czech Technical University in Prague \\
    Karlovo namesti 13, Prague 12135\\
  \texttt{rytirpav@fel.cvut.cz} \\
   \And
 Petr {Ry{\v{s}}av{\'y}} \\
 AI Center \\
    Czech Technical University in Prague \\
    Karlovo namesti 13, Prague 12135\\
  \texttt{petr.rysavy@fel.cvut.cz} \\
  \And
 Jakub Mare{\v {c}}ek \\ 
 AI Center \\
    Czech Technical University in Prague \\
    Karlovo namesti 13, Prague 12135\\
  \texttt{jakub.marecek@fel.cvut.cz} \\
}
\date{November 2026}

\begin{document}
\maketitle
\begin{abstract}

Exact score-based causal structure learning supports explicit structural constraints and optimisation certificates, but its branch-and-cut search can be computationally demanding. We augment exact causal structure learning methods with an upper bound on the maximum clique size of the learned undirected skeleton, enforced at integer candidates through lazy cycle and clique constraints. On static linear-Gaussian Erd\H{o}s--R\'enyi and scale-free DAGs, we evaluate callback order, alternative clique-cut policies, incumbent repair, fractional separation, and Big-$M$ bounds. We report causal-recovery metrics jointly with optimality gaps, primal and dual bounds, node throughput, MIPSOL events, and callback cost. Two variants receive particular attention: B3 selects the most violated minimal \((k+1)\)-clique inequality at each callback, whereas C3 repairs an infeasible integer candidate by deleting violating edges and refitting affected regression columns using QR-based least squares. The results do not indicate that clique constraints uniformly accelerate optimisation. However, their benefit depends on graph density, sample size, and cut-management policy, while large collections of correlated constraints can reduce node throughput. Tighter valid Big-$M$ bounds reduced the fixed-budget optimality gap in most tested settings, without necessarily changing SHD or F1 at a fixed time limit. These findings characterize the computational trade-offs of clique-constrained exact causal learning, and identify selective separation and incumbent repair as promising directions.
\end{abstract}

\section{Introduction}

Causal structure learning from observational data is a fundamental problem in
statistics and machine learning. Continuous methods such as NOTEARS represent
acyclicity through a differentiable equality constraint
\cite{zheng2018notears}. Exact score-based methods instead introduce binary
edge-selection variables and impose graph validity through integer or
mixed-integer constraints, thereby providing globally optimal or certifiably
near-optimal solutions under the specified model and stopping tolerance
\cite{cussens2014ipbn,bartlett2017Integer,cussens2020gobnilp}. Related exact
formulations apply this principle to causal structure learning under different
graphical assumptions, including extensions that represent latent confounding
\cite{exdag,exdbn,chen2021integer,exmag}. Although these formulations provide
a principled mechanism for encoding graph validity, sparsity, degree
restrictions, and prior knowledge, their computational cost remains
substantial: the number of admissible structures grows exponentially, and
solver performance depends on the formulation, relaxation strength, branching
behavior, and management of constraints generated during the search.

Graph inequalities and on-demand separation are established techniques in
combinatorial optimisation. Rather than enumerating every structural
constraint in the initial formulation, a branch-and-cut method identifies a
violated inequality at the current solution and adds it during the search
\cite{sonuc2015branch}. Following this principle, we impose an upper bound on
the clique size of the learned undirected skeleton. Unlike a global edge-count
or degree restriction, a maximum-clique constraint directly excludes locally
complete subgraphs and therefore serves as a local-density prior.

Enumerating all forbidden node subsets in advance is generally impractical, so
we enforce the maximum-clique condition lazily. At a MIPSOL event, the callback
examines an integer candidate and submits a lazy inequality through
\texttt{cbLazy()} when the candidate violates the prescribed clique bound.
This mechanism rejects structurally invalid integer candidates but does not
directly strengthen fractional node relaxations. Strengthening a fractional
relaxation instead requires a valid user cut submitted through
\texttt{cbCut()} at MIPNODE. The computational effect of lazy clique
separation is therefore empirical: a selected inequality may prevent repeated
integer violations, whereas separation and additional model rows may increase
callback overhead and node-relaxation cost.

The proposed condition should be interpreted specifically as a bound on the
maximum clique of the learned skeleton. It is not a bound on clique-width or
treewidth, and subsequent moralization or triangulation may introduce cliques
that are absent from the original skeleton. We consequently do not assume that
the restriction necessarily reduces downstream chordal complexity or
accelerates optimisation. Instead, we examine when the structural prior
improves causal-structure recovery or redirects the branch-and-cut search,
and when its additional separation cost outweighs the reduction in the
feasible graph class.

Our contributions
are fourfold. First, we
formulate and implement a lazy maximum-clique-size restriction for exact DAG
structure learning. Second, we compare callback order and controlled
clique-cut selection, including first and most-violated minimal-\((k+1)\)
policies. Third, we evaluate complementary
strategies involving incumbent repair, fractional separation, and tighter
Big-\(M\) bounds. Fourth, we introduce detailed instrumentation that separates
primal progress, dual-bound progress, MIPSOL events, node throughput, cut
submission, and callback time. These
measurements expose when clique constraints improve DAG recovery, when they
merely redirect the search, and when their additional optimisation cost
outweighs the reduction in the feasible graph class.

\section{Graph knowledge background}
An (undirected) graph $G = (V,E)$ consists of a set of nodes $V$ and a set of edges $E \subseteq V \times V$, where $(v_i,v_j) \in E$ indicates that nodes $v_i$ and $v_j$ are adjacent. For a graph $G$, we use $V(G)$ and $E(G)$ to denote its node set and edge set, respectively.

The adjacency matrix of a graph $G$ is denoted by $B_G$, where we put ones on its diagonal.
For two graphs $G$ and $H$, we say that $G$ is a subgraph of $H$ if
\[
V(G) \subseteq V(H) \quad \text{and} \quad E(G) \subseteq E(H),
\]
which is denoted by $G \subseteq H$.

For a graph $G=(V,E)$, a \emph{cycle of length $k$} is a sequence of distinct nodes
$v_1,v_2,\dots,v_k \in V$ such that $(v_i,v_{i+1}) \in E$ for $i=1,\dots,k-1$
and $(v_k,v_1) \in E$. 
A \emph{clique} $C \subseteq V$ of a graph $G = (V, E)$ is a subset of vertices where every two distinct nodes are adjacent, i.e., $(v_i, v_j) \in E$ for all $v_i, v_j \in C$ with $v_i \neq v_j$.

\section{Formulation}
We formulate causal structure learning as a mixed-integer optimisation problem. Let $n$ be the number of observations, $d$ the number of variables, $X \in \mathbb{R}^{n \times d}$ the
contemporaneous data matrix and $Y^{(t)} \in \mathbb{R}^{n \times d}$,
$t=1,\ldots,p$, the lagged data matrices. The optimisation variables are the contemporaneous weighted adjacency matrix $W \in \mathbb{R}^{d \times d}$, lagged coefficient matrices $A^{(t)}$, and binary edge-selection variables $z_{ij}\in\{0,1\}$ indicating whether edge $i\rightarrow j$ is selected. For each sample $i$ and node $j$, the residual is

\[
r_{ij}
=
X_{ij}
-
\sum_{k\neq j}X_{ik}W_{kj}
-
\sum_{t=1}^{p}\sum_{k=1}^{d}
Y_{ik}^{(t)}A_{kj}^{(t)}.
\]
This yields the following regularized least-squares objective:

\begin{equation}
\min_{W,A,z}
\;
\frac{1}{n}
\sum_{i=1}^{n}\sum_{j=1}^{d}r_{ij}^{2}
+
\frac{\lambda_1}{d}\,\mathrm{Reg}_W
+
\frac{\lambda_2}{d}\,\mathrm{Reg}_A,
\label{eq:milp-cliquewidth-objective}
\end{equation}
where
\[
\mathrm{Reg}_W=
\begin{cases}
\|W\|_F^2, & \text{if } \text{ Reg Type is $l_2$},\\
\|z\|_1, & \text{if } \text{ Reg Type is $l_1$},
\end{cases}
\]
\[
\mathrm{Reg}_A=
\begin{cases}
\displaystyle\sum_{t=1}^{p}\|A^{(t)}\|_F^2, & \text{if } \text{ Reg Type is $l_2$},\\
\displaystyle\sum_{t=1}^{p}\|A^{(t)}\|_1, & \text{if } \text{ Reg Type is $l_1$}.
\end{cases}
\]
With squared loss, the formulation is a mixed-integer quadratic program (MIQP). An MILP variant is obtained
with absolute loss by introducing auxiliary variables
$u_{ij}\ge0$, satisfying $-u_{ij}\le r_{ij}\le u_{ij}$ and replacing~\eqref{eq:milp-cliquewidth-objective} by

\[
\min_{W,A,z,u}
\;
\frac{1}{n}
\sum_{i=1}^{n}\sum_{j=1}^{d}u_{ij}
+
\frac{\lambda_1}{d}\,\mathrm{Reg}_W
+
\frac{\lambda_2}{d}\,\mathrm{Reg}_A,
\]
which linearizes the absolute residuals.

\paragraph{Big-$M$
constraints}
Binary variables are linked with edge weights through the standard big-$M$
constraints,
\begin{equation}
-Mz_{ij}\le W_{ij}\le Mz_{ij},
\qquad i\neq j,
\label{eq:bigM}
\end{equation}
where $M$ is a valid upper bound on the absolute edge weight. A smaller valid value strengthens the continuous relaxation, whereas an excessively large value permits a small fractional $z_{ij}$ to support a comparatively large $W_{ij}$. Reducing $M$ without validating the coefficient range can, however, exclude feasible solutions. To exclude two-node cycles, we impose
\begin{equation}
z_{ij}+z_{ji}\le1,
\qquad i\neq j.
\label{eq:antisymmetry}
\end{equation}
Longer directed cycles are eliminated lazily during branch-and-bound.
Whenever the callback detects a directed cycle $C$, it generates the cut
\begin{equation}
\sum_{(i,j)\in C}z_{ij}
\le
|C|-1,
\label{eq:cycle-cut}
\end{equation}
forcing at least one edge in every directed cycle to be removed.

\begin{figure*}[t]
\centering
\includegraphics[width=\textwidth]{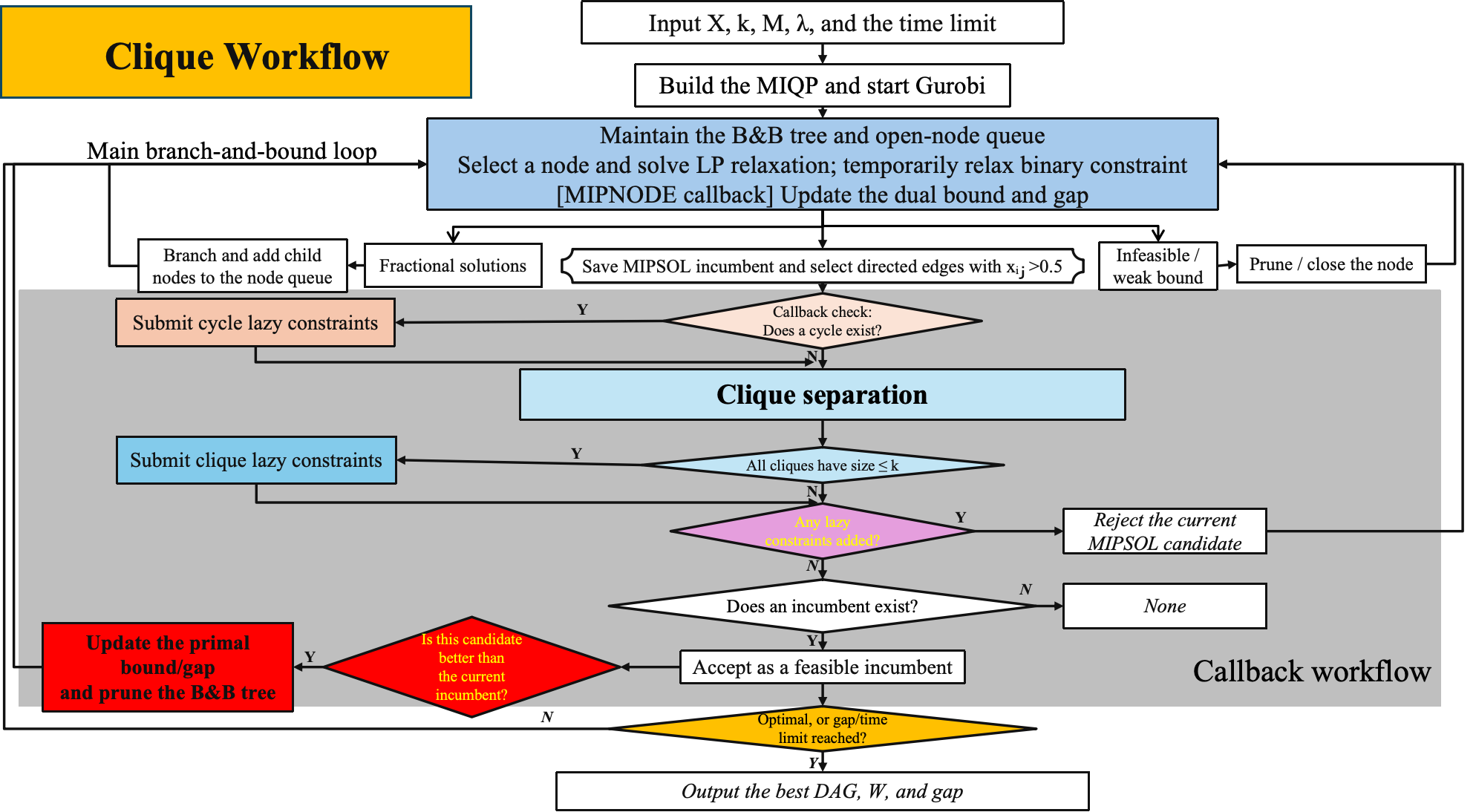}
\caption{\textbf{Workflow of clique-constrained exact causal structure learning.} The solver alternates between branch-and-cut node processing, cycle separation, clique separation, incumbent acceptance, and bound updates until optimality or a stopping limit is reached.}
\label{fig:clique-workflow}
\end{figure*}

\paragraph{Lazy clique constraints}
Let $Q$ be a clique in this skeleton such that $|Q| > k$, where $k$ is the maximum permitted
clique size (set to five in our experiments), and let
\[
P(Q) = \{(i,j) : i < j,\ i,j \in Q\}
\]
denote the set of unordered node pairs in $Q$. Since $|P(Q)|=\binom{|Q|}{2}$, the callback submits the lazy inequality
\begin{align}
\sum_{(i,j)\in P(Q)} (z_{ij}+z_{ji}) \le \binom{|Q|}{2}-1,
\label{eq:clique-cut}
\end{align}
which removes the violating clique while preserving all skeletons whose
maximum clique size is at most $k$.
The callback separates violated cliques during branch-and-cut. The complete
optimisation and callback workflow is summarized in
Figure~\ref{fig:clique-workflow}.

\section{Methodology}
\paragraph{Strength and selection of clique cuts.}
A limitation of Constraint~\eqref{eq:clique-cut} is that it only breaks the particular
complete clique detected by the callback; it does not force the remaining
graph to satisfy the maximum clique size $k$ immediately. For example, an
eight-node clique contains
\begin{equation}
|P(Q)|=\binom{8}{2}=28
\end{equation}
undirected edges, so the callback adds
\begin{equation}
\sum_{\{i,j\}\subseteq Q}(z_{ij}+z_{ji})\leq 27.
\label{eq:eight-clique-cut}
\end{equation}
This cut requires only one of the 28 edges to be removed. It excludes the
complete 8-clique, but the resulting graph may still contain a
7-clique or a 6-clique. When $k=5$, such candidates remain infeasible
even though they satisfy Constraint~\eqref{eq:eight-clique-cut}. The original
maximal-clique cut is therefore exact for the detected clique but weak with
respect to all forbidden subcliques, and several similar infeasible MIPSOL
candidates may be generated before repeated callback rounds eliminate them.

A stronger alternative separates minimal forbidden sets of size $k+1$. For
each $S\subseteq Q$ with $|S|=k+1$, the valid inequality is
\begin{equation}
\sum_{\{i,j\}\subseteq S}(z_{ij}+z_{ji})
\leq \binom{k+1}{2}-1.
\label{eq:minimal-clique-cut}
\end{equation}
For the 8-clique example with $k=5$, there are $\binom{8}{6}=28$ six-node subsets, each requiring at least one of its 15 skeleton edges to be absent. Adding every row excludes more related integer candidate at once, but can increase 
separation, model updating, and subsequent node-processing costs. We therefore evaluate policies that vary the number and diversity of submitted rows, including violation-based ranking, per call cut limits, duplicate suppression, Jaccard diversity, and search history~\cite{jaccard1912,huang2022cutselection,turner2023adaptive}.

\paragraph{B3: most-violated minimal clique cut.}
At a MIPSOL callback, B3 first detects each oversized clique $Q$ in the
integer candidate and generates its minimal forbidden subsets
$S\subseteq Q$, $|S|=k+1$. For each subset it evaluates the violation
\begin{align}
    v(S)=
\sum_{\{i,j\}\subseteq S}(z_{ij}+z_{ji})
-\left[\binom{k+1}{2}-1\right].
\label{eq:most-violated-cut}
\end{align}
The subsets are ranked by $v(S)$ and the highest-ranked row is submitted
through \texttt{cbLazy()} (subject to the configured per-callback limit).
B3 is consequently an integer-candidate cut-selection policy, not a MIPNODE
fractional separator. At a binary clique candidate, multiple minimal subsets
can have identical violation; in that case B3 reduces to a deterministic
tie-breaking rule and may behave similarly to the first-\((k+1)\) policy in method B1 (included in the Appendix). The Clique-Most-Violated (B3) implementation proceeds
as follows.
\begin{enumerate}[leftmargin=3em]
    \item[\textbf{Step 1.}] Read the integer edge-selection candidate and
    detect every maximal clique \(Q\) whose size exceeds \(k\).

    \item[\textbf{Step 2.}] Generate the unique minimal forbidden node sets
    \(S\subseteq Q\) with \(|S|=k+1\). If several maximal cliques contain the
    same \(S\), that subset appears only once in the candidate list.

    \item[\textbf{Step 3.}] For every candidate \(S\), construct its unordered
    node pairs and evaluate in Equation~\ref{eq:most-violated-cut}.
    The implementation also computes the normalized efficiency
    \(v(S)/\binom{k+1}{2}\).

    \item[\textbf{Step 4.}] Rank candidates lexicographically by decreasing
    violation, decreasing efficiency, and then increasing sorted node tuple.
    The node tuple therefore provides deterministic tie-breaking.

    \item[\textbf{Step 5.}] Select the highest-ranked row and apply the
    configured callback cut limit. In the B3 experiments,
    \texttt{max\_clique\_cuts\_per\_callback=1} and clique-cut
    deduplication is disabled, so at most one row is submitted during the
    callback.

    \item[\textbf{Step 6.}] Submit the selected inequality through
    \texttt{cbLazy()}, reject the current structurally invalid integer
    candidate, return control to Gurobi, and resume branch-and-cut search.
\end{enumerate}
\paragraph{C3: QR-based incumbent repair.}
C3 supplements lazy rejection with a primal heuristic. When a MIPSOL candidate
violates acyclicity or the clique bound, it retains the candidate support and
weights initially, greedily deletes edges implicated in the violations, and
records the parent sets that changed. Only regression columns associated with
those parent sets are refitted by QR-based least squares. Before submission,
the repaired assignment is validated against integrality, Big-\(M\) linking,
acyclicity, and the maximum-clique constraint. A valid repair is submitted with
\texttt{cbSetSolution()} and evaluated with \texttt{cbUseSolution()}, while
the original infeasible candidate remains rejected through the required lazy
constraints. C3 can improve the primal search by reusing an otherwise rejected
candidate, but it neither adds fractional user cuts nor directly strengthens
the dual bound.

\section{Experiments}
This section describes the data-generation procedure, solver configuration, experimental design, and empirical results.

\subsection{Synthetic Data Generation}
We use synthetic data because they provide a known ground-truth adjacency matrix. The benchmark contains clique-filtered scale-free (cliquefree-SF) and Erd\H{o}s--R\'enyi (cliquefree-ER) DAGs in a static linear-Gaussian setting ($p=0$), following the general simulation protocol of NOTEARS~\cite{zheng2018notears}. The SF generator produces sparse, hub-dominated graphs. For ER data, an undirected graph is sampled at the target edge ratio and oriented according to a random topological ordering. In both families, oversized cliques are detected and low-weight edges are removed until the prescribed clique bound is satisfied. Observations are then sampled from the resulting linear structural equation model (SEM).

Let $W\in\mathbb{R}^{d\times d}$ denote the weighted adjacency matrix of the ground-truth DAG, where $W_{ii}=0$ and $W_{ij}\neq 0$ if and only if the edge $i\to j$ is present in the data-generating graph. A sample $x\in\mathbb{R}^{d}$ is generated according to
\begin{align}
 x = W^\top x + \varepsilon,
 \qquad
 \varepsilon \sim \mathcal{N}(0, I_d).
 \label{eq:sem_static}
\end{align}
Because the ground-truth graph is acyclic, $I-W^\top$ is invertible. 

Samples are generated in topological order, with nonzero coefficients drawn from the configured weight intervals and unit-variance Gaussian noise. We consider $d\in\{11,15,20,25,30,35,40,45,50\}$ nodes and $n\in\{200,2000\}$ observations. The nominal ER edge ratio is $4$ or $5$. Because the reported experiments are static, the inter-slice edge-ratio parameter is inactive. Detailed density and linearity statistics are reported in the supplementary material.

\subsection{Solver Configuration}
We use Gurobi with lazy cycle separation enabled. Unless otherwise stated, each run has a 1,800-second time limit, a target relative MIP gap of $10^{-3}$, $M=100$, squared loss, and the regularization parameters $\lambda_1=\lambda_2=1$. Dedicated Big-\(M\) experiments evaluate alternative valid bounds. Hydra manages configurations and MLflow records parameters, metrics, and run metadata~\cite{Yadan2019Hydra,zaharia2020mlflow}. Experiments were executed on MetaCentrum~\cite{sitera2011metacentrum} using jobs allocated two CPUs, 32~GB RAM, and 30~GB of local scratch storage. The software environment uses Python~3.10, Gurobi, NetworkX, NumPy, and SciPy. Detailed implementation and environment information are provided in the supplementary code-and-data package.

\subsection{Experiment Setup}
Table~\ref{tab:method-variants} summarizes ExDBN, Clique, and the principal variants evaluated on static cliquefree-SF and cliquefree-ER data. For each graph family and each $(d,n)$ pair, we use four seeds ($1,2,3,4$). We record structural Hamming distance (SHD), F1 score, the relative MIP gap, submitted lazy constraints, primal and dual bounds, and solver diagnostics. We evaluate the best incumbent available at time limits of $5,60,120,300,600,1200,1800,2700,3600,5400,$ and $7200$ seconds. Thus, each fixed-time comparison concerns the best solution found within the specified budget rather than a final globally optimal solution.
\begin{table*}[t]
\centering
\small
\setlength{\tabcolsep}{5pt}
\renewcommand{\arraystretch}{1.12}
\begin{tabular}{
    p{3.0cm}
    p{5.2cm}
    p{5.0cm}
    >{\centering\arraybackslash}p{1.5cm}
}
\toprule
\textbf{Method} &
\textbf{Integer-separation policy} &
\textbf{Repair} &
\textbf{$M$} \\
\midrule

\textbf{ExDBN} &
None: the baseline enforces acyclicity but imposes no clique-size restriction. &
None &
100 \\
\addlinespace

\textbf{Clique} &
At each infeasible integer candidate, submit lazy cycle constraints and all detected violated clique constraints. &
None &
100 \\
\addlinespace

\textbf{Clique $\cdot$ Most-Violated} &
At each infeasible integer candidate, submit lazy cycle constraints and only the most-violated minimal $(k+1)$ clique constraint. &
None &
100 \\
\addlinespace

\textbf{Clique $\cdot$ QR Repair} &
Use the default Clique integer separation: submit lazy cycle and violated clique constraints. &
Delete violation-causing edges, locally refit affected regressions by QR, validate the repaired assignment, and submit it via \texttt{cbSetSolution()}. &
100 \\
\addlinespace

\textbf{Clique $\cdot$ $M=20$} &
Use the default Clique integer separation. &
None &
20 \\
\bottomrule
\end{tabular}
\caption{Compared methods. Integer separation is performed at \texttt{MIPSOL} events through lazy constraints; fractional \texttt{MIPNODE} separation is evaluated separately.}
\label{tab:method-variants}
\end{table*}

\begin{figure*}[t]
\centering
\includegraphics[width=0.48\textwidth]{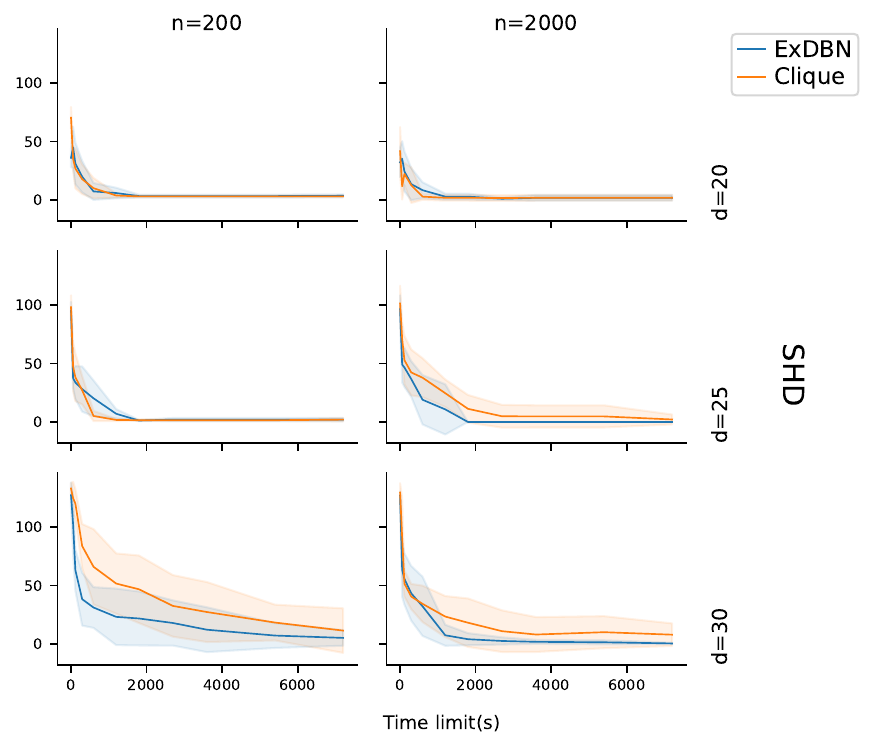}
\includegraphics[width=0.48\textwidth]{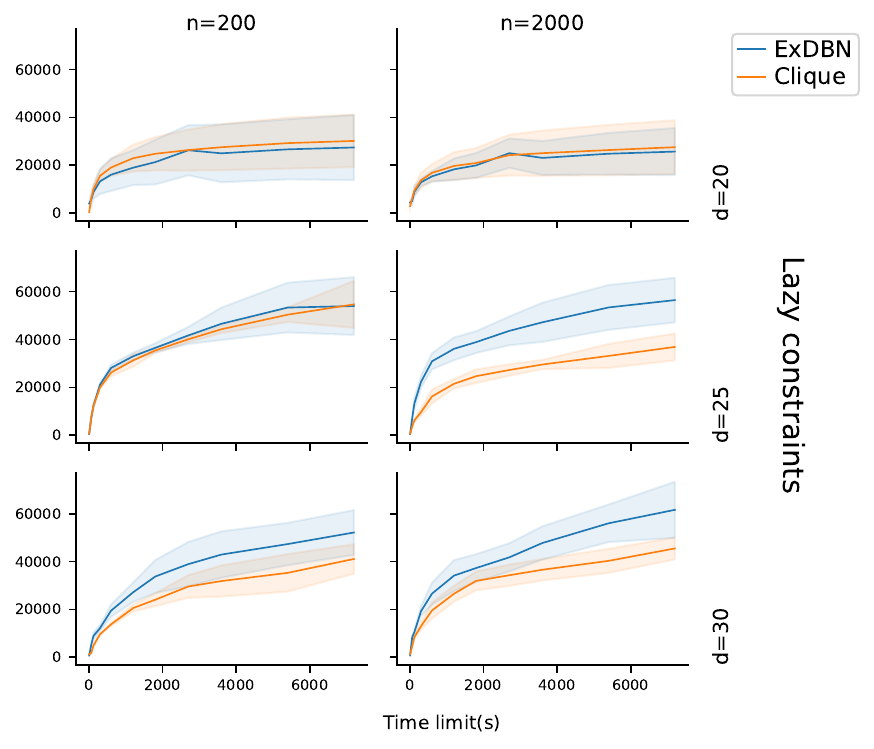}
\caption{\textbf{Comparison of the SHD score and number of lazy-constraint submissions between Clique and ExDBN under different time limits setting and various dimensions $d$.} The solid line represents the mean across multiple trials, with the shaded region indicating the standard deviation.}
\label{fig:results_timelimit_lazy_count_and_SHD}
\end{figure*}

\subsection{Observations}
Most large instances terminate at the time limit with a nonzero optimality gap. Accordingly, every fixed-time comparison concerns the best incumbent and best bound available at that budget, not a proven global optimum. A higher SHD for Clique at a fixed time therefore indicates weaker time-limited recovery on that run; it does not establish that the structural constraint reduces the accuracy of the fully optimized model. Under the fixed $1000$-second computational budget, the reported gaps and node throughputs reflect solver progress rather than certified final-optimum quality.

SHD and F1 evolve only when a newly accepted incumbent changes the learned graph. They can therefore remain constant while the solver continues to improve the dual bound and close the MIP gap. This stepwise behavior is visible in the time-limit experiments: recovery metrics often stabilize before the optimality certificate. It would nevertheless be incorrect to infer a monotone relation between gap and SHD. A smaller gap certifies the current objective more tightly, whereas SHD measures agreement with an external ground-truth graph; either quantity can improve without the other.

Thus, the poorer SHD of Clique in Figure~\ref{fig:results_timelimit_lazy_count_and_SHD} cannot be directly interpreted as “the constraints reduce accuracy.” A more accurate statement is that, under the same time budget, Clique has not yet found an incumbent of the same quality as ExDBN. For \(d=20\), after 1,200 seconds, the gaps of the Clique and ExDBN optimisation models continue to decrease gradually, while the SHD score (and F1 score) of the predicted causal graphs tend to stabilize. For \(d=25\), the overall change in SHD slows after 2,700 seconds. For \(d=30\), the overall change in SHD score slows after 3,000 seconds.

The distance of the corresponding graphical causal-model solution gradually decreases, but the numbers of callbacks are different when the solutions reach a stable state. For \(d=30\), after the optimisation process reaches a stable regime, the stable number of lazy callbacks for ExDBN is larger than that for Clique, while their SHDs are similar.
It cannot be concluded that ExDBN produces a more accurate causal model. Because Clique introduces violation checks for cliques larger than $5$, when the clique and cycle violation checks do not overlap, Clique should have more callbacks than ExDBN. One plausible explanation, consistent with the callback diagnostics, is adding clique callbacks reduces the number of cycle callbacks, and the number of added clique callbacks is smaller than the corresponding reduction in cycle callbacks.

\begin{figure}[t]
\centering
\includegraphics[width=0.52\textwidth]{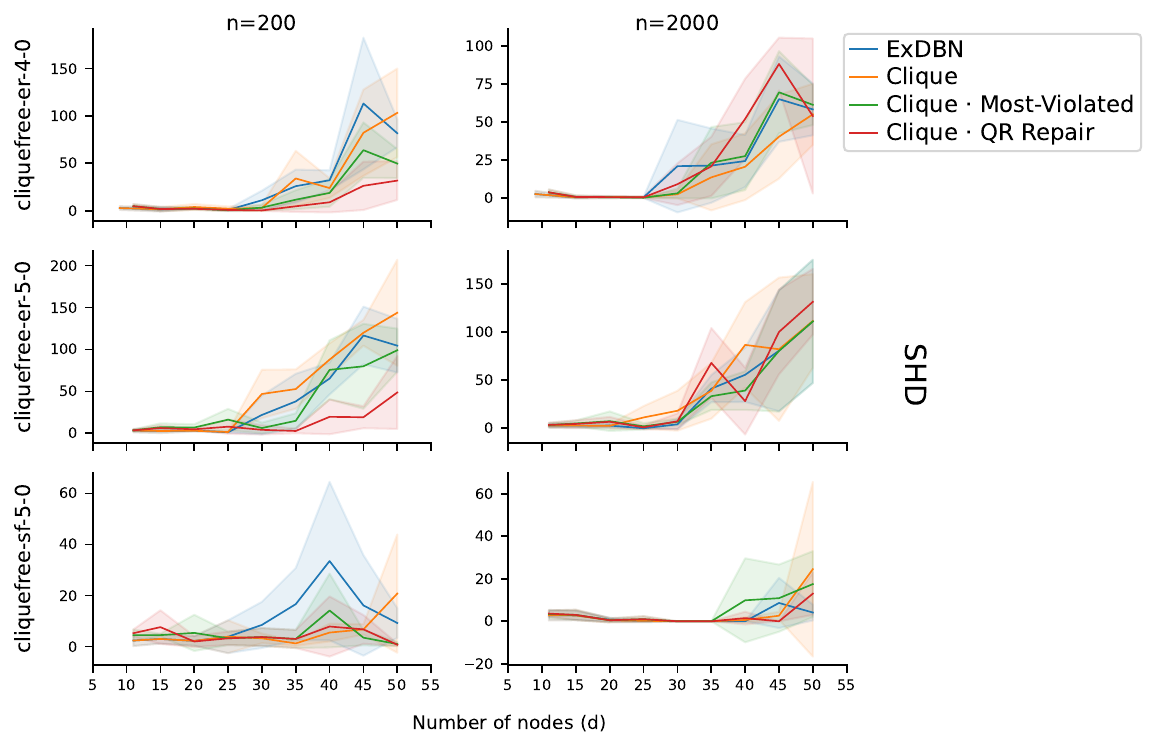}
\caption{\textbf{SHD results over the ER and scale-free synthetic graph families.} Solid lines and shaded areas denote the mean and standard deviation of multiple runs, respectively.}
\label{fig:results_nodes}
\end{figure}
Figure~\ref{fig:results_nodes} shows the Structural Hamming Distance (SHD) of four different methods as the number of nodes ($d$) increases from 5 to 55. 
In the \textit{cliquefree-er-4-0} and \textit{cliquefree-er-5-0} graph settings, the SHD of all methods remains low when $d \leq 30$, but increases when $d > 35$. 
Among them, \textit{Clique $\cdot$ QR Repair} generally results in a lower SHD than \textit{ExDBN} and standard \textit{Clique} in these two settings when the graph scale grows larger. 
For the \textit{cliquefree-sf-5-0} setting, the SHD values of all four methods stay below 50 across all tested values of $d$, which is lower than the values observed in the other two graph types. 
For $n=200$, Clique · QR Repair attains the lowest SHD on both ER settings at $d\ge 40$; for $n=2000$, this advantage disappears and the methods are not consistently ordered.

\begin{figure}[t]
\centering
\includegraphics[width=0.48\textwidth]{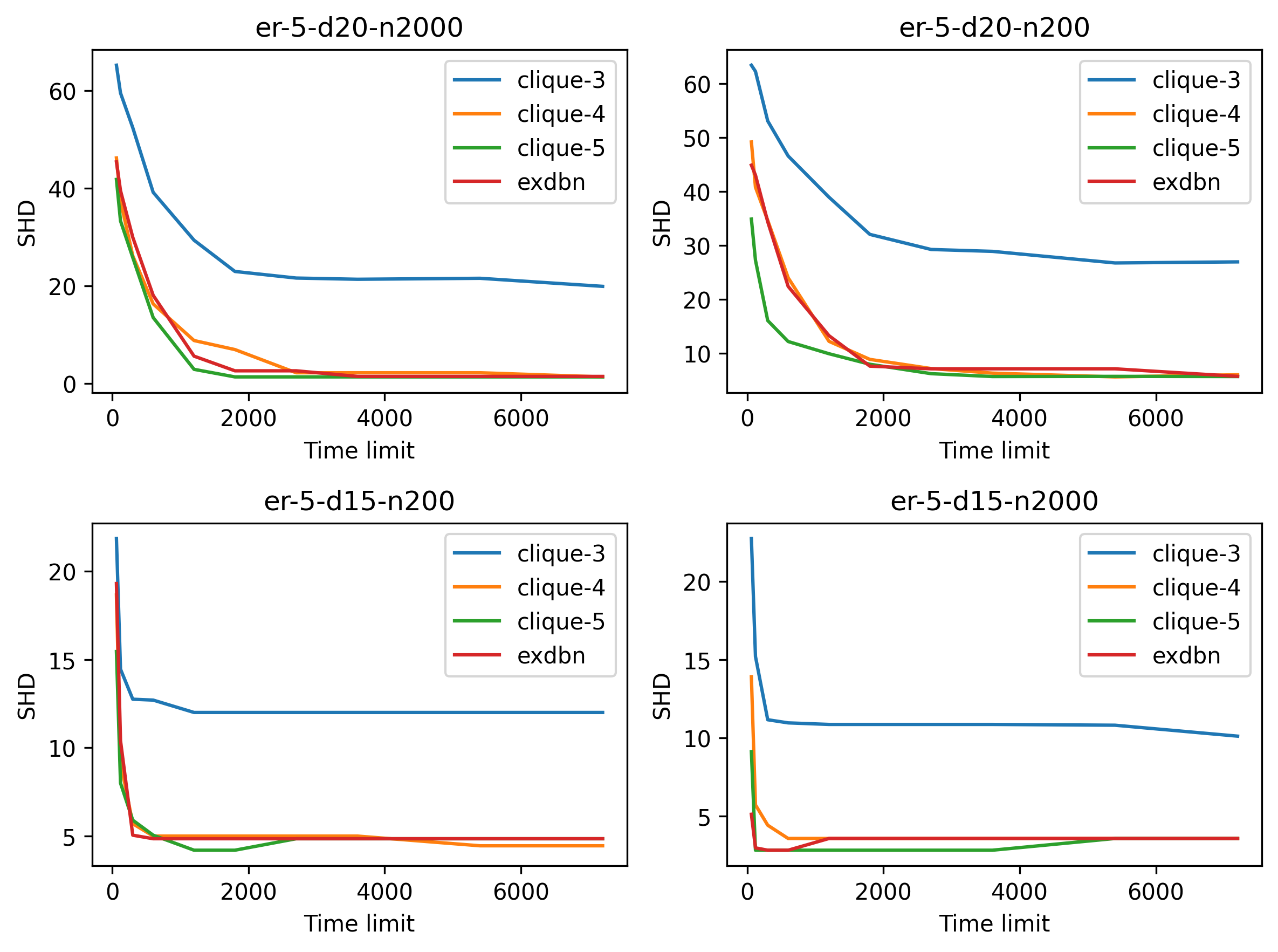}
\caption{\textbf{Comparison between ExDBN and Cliques with maximum clique sizes
$k\in\{3,4,5\}$ under different time limit settings.} }
\label{fig:results_clique_size_shd}
\end{figure}

In Figure~\ref{fig:results_clique_size_shd}, clique-5 provides the most stable performance, while the optimal clique bound remains dependent on graph size and sample size.

\paragraph{Big-$M$ and fractional separation.}
Table~\ref{tab:bigm-fractional} reports fixed-budget optimisation
diagnostics. Under the $1{,}000$-second budget, the effect of the linking
bound is dataset-dependent. For ER-4-0, $M=20$ achieved the smallest mean
optimality gap ($10.03\%$, compared with $17.60\%$ for $M=100$) and the
highest node throughput. This advantage did not transfer uniformly to
ER-5-0: $M=10$ obtained the smallest gap ($11.05\%$), while the results
for $M=20$ and $M=100$ were close ($11.39\%$ and $11.65\%$,
respectively). Thus, the results support using a tighter \emph{valid}
linking bound, but do not identify $M=20$ as uniformly optimal across
instances.

\begin{figure*}[t]
\centering
\includegraphics[width=\textwidth]{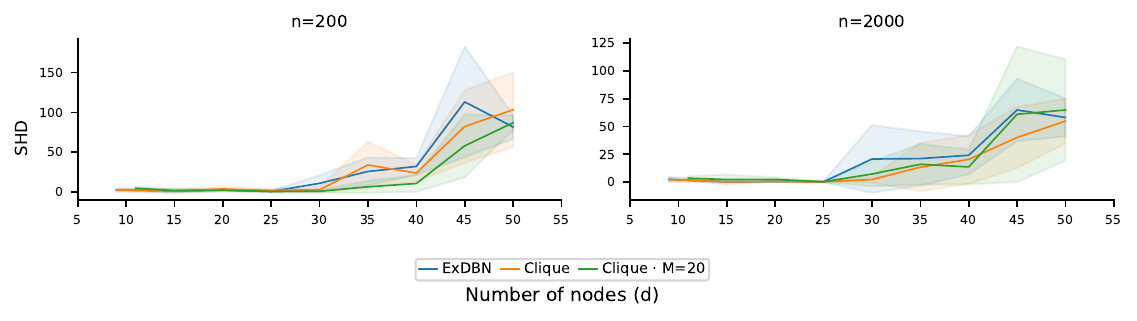}
\caption{\textbf{SHD on ER-4-0 as the number of variables increases.}
Clique with $M=20$ is competitive with ExDBN and the default Clique
formulation, but its advantage is not uniform across node counts or sample
sizes. Shaded regions show variability across seeds.}
\label{fig:bigm-shd}
\end{figure*}

Figure~\ref{fig:bigm-shd} complements these optimisation diagnostics with
graph-recovery results on ER-4-0 over increasing numbers of variables.
The SHD curves show that Clique with $M=20$ is generally competitive with
the default Clique formulation and ExDBN, particularly at larger problem
sizes; however, the uncertainty intervals widen substantially as the
number of variables increases, and no method dominates uniformly for both
sample sizes. Therefore, the improvement observed for $M=20$ in the
fixed-budget ER-4-0 optimisation experiment should not be interpreted as
a universal improvement in structural-recovery accuracy.

Fractional clique separation did not improve performance in the current
implementation. The root-exhaustive, root-greedy, and history-guided
variants (D2, D3, and D5) submitted no user cuts in these runs and
therefore remained close to the no-fractional-cut baseline (D1), as shown
in Table~\ref{tab:bigm-fractional}. The sparse non-root variant (D4)
submitted, on average, 146.5 and 108.5 user cuts on ER-4-0 and ER-5-0,
respectively; however, callback time increased to $41.09\%$ and
$47.38\%$ of total runtime, while node throughput decreased substantially.
In particular, its ER-5-0 gap increased from $11.66\%$ for D1 to
$12.82\%$ for D4. These results do not support enabling the current
fractional-separation procedure by default.

\begin{table}[t]
\centering
\small
\setlength{\tabcolsep}{2.2pt}
\renewcommand{\arraystretch}{1.08}
\begin{tabular}{@{}p{1.35cm}p{2.15cm}rr@{}}
\toprule
\textbf{Study} & \textbf{Setting} &
\textbf{ER-4-0} & \textbf{ER-5-0} \\
\midrule

Big-$M$
& $M=10$  & 16.75\% / 2147 & 11.05\% / 2733 \\
& $M=20$  & \textbf{10.03\% / 2621} & \textbf{11.39\% / 2779} \\
& $M=50$  & 13.81\% / 2412 & 12.03\% / 2755 \\
& $M=100$ & 17.60\% / 2453 & 11.65\% / 2774 \\
\midrule

\multirow{5}{*}{\shortstack[l]{Fractional\\separation}}
& No \texttt{cbCut}
& 20.20\% / 2110 & \textbf{11.66\% / 2760} \\

& Root exhaustive
& 20.24\% / 2096 & 11.69\% / 2718 \\

& Root greedy
& 20.27\% / 2088 & 11.69\% / 2721 \\

& \shortstack[l]{Root + sparse\\non-root}
& \textbf{20.18\% / 1297} & 12.82\% / 1330 \\

& History-guided
& 20.53\% / 2001 & 11.79\% / 2582 \\
\bottomrule
\end{tabular}

\caption{\textbf{Big-$M$ and fractional-separation ablation.}
Each cell reports Gap (\%) / nodes per second under the fixed
$1{,}000$-second budget. D2, D3, and D5 submitted no user cuts. D4 submitted
146.5 / 108.5 user cuts on ER-4-0 / ER-5-0, respectively, while spending
41.09\% / 47.38\% of runtime in callbacks.}
\label{tab:bigm-fractional}
\end{table}

\paragraph{Callback-order comparison.}
Table~\ref{tab:callback-order-statistics} compares Cycle-only,
Cycle-then-Clique, and Clique-then-Cycle under a fixed 1,800-second
budget. Since the remaining MIP gaps are large, the reported quantities
reflect intermediate solver progress rather than certified optimality.

For ER-4-0, the mixed policies explore substantially fewer nodes than
Cycle-only (348{,}106 and 379{,}203 versus 1{,}914{,}601). Both mixed
policies also submit fewer lazy constraints on ER-4-0 and ER-5-0; when
cycles are separated first, the later clique stage submits relatively few
clique constraints. However, this search-tree reduction does not improve
the fixed-budget Gap: Cycle-only has the smallest Gap and largest best
bound on both datasets. Thus, mixed callback orders reduce constraint
submissions and may reduce node exploration, but do not uniformly improve dual-bound progress in the current implementation.
\begin{table}[t]
\centering
\footnotesize
\setlength{\tabcolsep}{2.2pt}
\renewcommand{\arraystretch}{1.04}
\begin{tabular}{llrrr}
\toprule
\textbf{Ratio} & \textbf{Metric} &
\textbf{Cycle-only} &
\textbf{Cycle$\rightarrow$Clique} &
\textbf{Clique$\rightarrow$Cycle} \\
\midrule

\multirow{5}{*}{4}
& Submitted ($10^3$) & 23.72 & 18.42 & 20.16 \\
& Nodes ($10^6$)     & 1.915 & 0.348 & 0.379 \\
& Gap (\%)           & \textbf{52.81} & 59.69 & 59.91 \\
& Best bound         & \textbf{11.234} & 9.600 & 9.545 \\
& MIPSOL events      & 532 & \textbf{398} & 433 \\
\midrule

\multirow{5}{*}{5}
& Submitted ($10^3$) & 31.18 & \textbf{29.23} & 29.56 \\
& Nodes ($10^6$)     & 1.586 & 1.128 & \textbf{1.014} \\
& Gap (\%)           & \textbf{53.68} & 55.32 & 56.29 \\
& Best bound         & \textbf{11.336} & 11.017 & 10.779 \\
& MIPSOL events      & 546 & \textbf{529} & 580 \\
\bottomrule
\end{tabular}

\caption{\textbf{Callback-order diagnostics under a fixed budget.} All experiments use static Erd\H{o}s--R\'enyi synthetic data with $p=0$ and edge ratios of 4 and 5. Results are
averages over seeds 1--4 with $d=20$, $n=2000$, $M=100$, and a
1,800-second time limit. Submitted constraints and explored nodes are
reported in thousands and millions, respectively.}
\label{tab:callback-order-statistics}
\end{table}

\section{Conclusion}
We studied an explicit maximum-clique-size restriction for exact causal structure learning in the ExDBN mixed-integer formulation. Lazy constraints exclude violating integer candidates without enumerating all forbidden
subsets in advance. When the solver terminates with a nonzero gap, the returned incumbent is feasible under the generated structural constraints but is not claimed to be globally optimal.

The experiments show that clique constraints introduce a trade-off rather than a uniform speedup. Selective minimal cuts can reduce redundant model growth, but integer violation scores often tie, which limits the distinction between B3 and a deterministic first-\((k+1)\) rule. C3 reuses rejected candidates through QR-based repair and can improve primal search on some datasets, but its benefit is not consistent across all graph families. Fractional user cuts must be evaluated separately from MIPSOL lazy cuts because only the former directly strengthen node relaxations.

Finally, the Big-\(M\) results support tighter validated coefficient bounds: relative to $M=100$, smaller valid values generally improve relaxation quality and dual-bound progress, although the best tested value depends on the dataset. Future work should validate edge-specific bounds, evaluate B3 with fractional-aware scores, and report repair acceptance and incumbent-improvement rates for C3. These conclusions are limited to synthetic linear-Gaussian instances and runs that frequently terminate before proven optimality; evaluation on real, nonlinear, and larger-scale data remains necessary.

\section*{Acknowledgements}
This work was supported by the Ministry of Education, Youth and Sports of the Czech Republic through the e-INFRA CZ (ID:90254) project.

\bibliographystyle{unsrt} 
\bibliography{refs}

\newpage
\renewcommand{\thefigure}{S\arabic{figure}}
\renewcommand{\theequation}{S\arabic{equation}}
\renewcommand{\thetable}{S\arabic{table}}
\appendix

This supplementary appendix extends the main paper ``Clique Constraints for Exact Causal Structure Learning'' by providing detailed method explanations, technical clarifications, and additional empirical results. Specifically, we evaluate synthetic graph densities and linear $R^2$
, dissect the diagnostic timeline while defining the core \texttt{MIPSOL} and \texttt{MIPNODE} callback quantities, catalog the experiments A-G based on improving directions and attach a table of mathematical notations and optimization terminology
. Collectively, these materials serve as diagnostic support for the theoretical and empirical claims established in the main manuscript.

\section{Synthetic Data Density and Linearity}
\label{sec:data}

\begin{table*}[t]
\centering
\scriptsize
\renewcommand{\arraystretch}{1.08}
\setlength{\tabcolsep}{3pt}
\resizebox{\textwidth}{!}{%
\begin{tabular}{cccccccccccc}
\toprule
\multicolumn{4}{c}{\textbf{ER-4-0}} &
\multicolumn{4}{c}{\textbf{ER-5-0}} &
\multicolumn{4}{c}{\textbf{SF-5-0}} \\
\cmidrule(lr){1-4}\cmidrule(lr){5-8}\cmidrule(lr){9-12}
\textbf{Nodes} & \textbf{Nom.} & \textbf{Real.} & $\boldsymbol{R^2}$ &
\textbf{Nodes} & \textbf{Nom.} & \textbf{Real.} & $\boldsymbol{R^2}$ &
\textbf{Nodes} & \textbf{Nom.} & \textbf{Real.} & $\boldsymbol{R^2}$ \\
\midrule

11 & 80.00\% & 75.45\% / 77.27\% & 0.847 / 0.875
& 11 & 100.00\% & 57.58\% / 58.18\% & 0.737 / 0.808
& 11 & 100.00\% & 70.18\% / 70.38\% & 0.890 / 0.884 \\

15 & 57.14\% & 56.90\% / 56.90\% & 0.928 / 0.923
& 15 & 71.43\% & 62.82\% / 63.33\% & 0.897 / 0.916
& 15 & 71.43\% & 55.57\% / 55.60\% & 0.875 / 0.874 \\

20 & 42.11\% & 41.97\% / 42.01\% & 0.900 / 0.897
& 20 & 52.63\% & 48.31\% / 49.25\% & 0.907 / 0.926
& 20 & 52.63\% & 43.87\% / 42.63\% & 0.874 / 0.879 \\

25 & 33.33\% & 33.33\% / 33.33\% & 0.910 / 0.913
& 25 & 41.67\% & 41.16\% / 41.16\% & 0.929 / 0.930
& 25 & 41.67\% & 35.05\% / 35.28\% & 0.866 / 0.865 \\

30 & 27.59\% & 27.59\% / 27.59\% & 0.909 / 0.908
& 30 & 34.48\% & 34.03\% / 33.93\% & 0.918 / 0.921
& 30 & 34.48\% & 29.90\% / 30.06\% & 0.875 / 0.881 \\

35 & 23.53\% & 23.53\% / 23.53\% & 0.927 / 0.925
& 35 & 29.41\% & 29.41\% / 29.41\% & 0.941 / 0.941
& 35 & 29.41\% & 26.06\% / 26.11\% & 0.871 / 0.874 \\

40 & 20.51\% & 20.51\% / 20.51\% & 0.902 / 0.902
& 40 & 25.64\% & 25.64\% / 25.64\% & 0.951 / 0.951
& 40 & 25.64\% & 23.03\% / 23.12\% & 0.861 / 0.864 \\

45 & 18.18\% & 18.18\% / 18.18\% & 0.929 / 0.930
& 45 & 22.73\% & 22.73\% / 22.73\% & 0.943 / 0.943
& 45 & 22.73\% & 20.66\% / 20.73\% & 0.866 / 0.866 \\

50 & 16.33\% & 16.33\% / 16.33\% & 0.924 / 0.926
& 50 & 20.41\% & 20.41\% / 20.41\% & 0.956 / 0.957
& 50 & 20.41\% & 18.70\% / 18.68\% & 0.887 / 0.875 \\
\bottomrule
\end{tabular}
}
\caption{\textbf{Synthetic-data density and linearity.}
\textbf{Nom.} is the nominal skeleton density, and \textbf{Real.} is the
mean realized skeleton density after clique-free filtering.
Each entry is reported as $n=200$ / $n=2000$.
All values are means across the available generated instances.}
\label{tab:dataset-characteristics}
\end{table*}

\begin{table}[h]
\centering
\small
\renewcommand{\arraystretch}{1.1}
\setlength{\tabcolsep}{8pt}
\begin{tabular}{c c c c}
\toprule
\textbf{Edge Ratio} & \textbf{$d_{\max}$} & \textbf{Target Edges} & \textbf{Max Capacity} \\
\midrule
2 & 5  & 10 & 10 \\
3 & 7  & 21 & 21 \\
4 & 9  & 36 & 36 \\
5 & 11 & 55 & 55 \\
\bottomrule
\end{tabular}
\caption{\textbf{Theoretical limits for complete network saturation.} The maximum node number $d_{\max}$ represents the upper bound where an edge ratio $r$ can force a $100\%$ nominal skeleton density.}
\label{tab:saturation-limits}
\end{table}

In the main results, we observed that the clique-constrained callback enhances search efficiency more significantly in denser causal graphs. To dig deeper into the performance of these constraints across different scenarios, we examine three key dataset characteristics: nominal skeleton density, realized skeleton density, and linear $R^2$. The nominal skeleton density is calculated during the initial random graph generation step and remains invariant across random seeds. Conversely, the realized density represents the ratio of surviving edges in the true adjacency matrix $W_{\text{true}}$ to the maximum possible skeleton capacity after applying the clique-free filter (which removes cliques larger than $k$). These metrics are formalized as follows:
\begin{equation}
    \rho_{\text{nom}} = \min \left( 1, \frac{2r}{d - 1} \right), \quad \rho_{\text{real}} = \frac{|E(W_{\text{true}})|}{\binom{d}{2}},
\end{equation} 
where $r$ denotes the target edge ratio and $d$ represents the number of nodes.

As reported in Table~\ref{tab:dataset-characteristics}, while \texttt{ER-5-0} generates denser causal graphs than \texttt{ER-4-0} prior to filtering, its realized skeleton density drops below that of \texttt{ER-4-0} at $d = 11$. This structural inversion occurs because a higher initial edge ratio paired with a low node count severely exacerbates the formation of large, dense cliques; consequently, the clique-free filter must aggressively remove a substantially larger volume of violating edges to satisfy the structural constraint. This phenomenon highlights why comparative analyses should rely on the realized density rather than the nominal edge-ratio label alone. Complementing this finding, a comparison between the realized skeleton densities of \texttt{ER-5-0} and \texttt{SF-5-0} highlights the structural impact of their underlying generation algorithms. Specifically, unlike the random graph model, the scale-free model (\texttt{SF-5-0}) does not exhibit an unusually depressed realized density at $\text{Nodes} = 11$. This stability arises because its power-law degree distribution naturally distributes edges in a way that minimizes the dense, localized clique clusters typically found in Erd\H{o}s-R\'enyi configurations. Consequently, structural evaluations must distinguish between different graph typologies rather than relying solely on the edge-ratio baseline.

This structural behavior at low dimensions is directly linked to network saturation boundaries. As established previously, for a given edge ratio $r$, a graph achieves full saturation—reaching a $100\%$ nominal skeleton density where the target edge count equals or exceeds the total capacity of a complete graph, i.e.,
\begin{equation}
    \lfloor r \cdot d \rfloor \ge \binom{d}{2}
\end{equation}
—only when the node count $d$ is sufficiently small. Because $\lfloor r \cdot d \rfloor \le r \cdot d$, this condition implies $r \cdot d \ge \frac{d(d-1)}{2}$, which analytically simplifies to the deterministic boundary $d \le 2r + 1$. As $d$ grows beyond this threshold, the quadratic capacity $\frac{d(d-1)}{2}$ quickly outpaces the linear growth of the target edge allocation. The exact maximum node numbers ($d_{\max}$) for which each edge ratio completely saturates a graph are summarized in Table~\ref{tab:saturation-limits}. Given these boundary constraints, cross-model performance comparisons should avoid configurations starting at or below these minimal saturation thresholds to ensure fair evaluation.

Additionally, to quantify the degree of linearity and data-generating fidelity of our synthetic graphs, we evaluate the mean linear $R^2$ across all variables. For a specific node $j \in V$, let $X_j$ denote its observed data vector and $\textbf{Pa}(j)$ represent the set of its true structural parents determined by $W_{\text{true}}$. The linear $R^2$ for node $j$ is formalized as:
\begin{equation}
    R^2_j = 1 - \frac{\sum_{i=1}^n \left( X_{ij} - \hat{X}_{ij} \right)^2}{\sum_{i=1}^n \left( X_{ij} - \bar{X}_j \right)^2} = 1 - \frac{\text{RSS}_j}{\text{TSS}_j},
\end{equation}
where $\hat{X}_{ij}$ is the value predicted by an ordinary least squares (OLS) linear regression of $X_j$ on $X_{\textbf{Pa}(j)}$, $\bar{X}_j$ is the empirical mean of $X_j$, $\text{RSS}_j$ is the residual sum of squares, and $\text{TSS}_j$ is the total sum of squares. 

The globally reported linear $R^2$ profile in Table~\ref{tab:dataset-characteristics} represents the macro-average across all non-root nodes with non-empty parent sets, computed as:
\begin{equation}
    R^2_{\text{mean}} = \frac{1}{|V_{\text{non-root}}|} \sum_{j \in V_{\text{non-root}}} R^2_j.
\end{equation}
The linear $R^2$ values, which strictly range between $0$ and $1$, directly quantify the capability of true structural parents to linearly explain the variance of their corresponding child variables. In our experiments, the consistently high baseline ($R^2 \ge 0.85$, with many configurations exceeding $0.90$) demonstrates that a vast majority of the variable variance is fully accounted for by linear combinations of true causal ancestors, signifying a robust causal signal-to-noise ratio with minimal noise interference. Conversely, an exceptionally depressed $R^2$ (e.g., $< 0.40$) would imply either a dominance of random error or severe underlying non-linearities, under which standard linear Mixed-Integer Programming (MIP) optimization models would encounter immense difficulty in accurately recovering the true graph topology.

When scaling the sample size from $n=200$ to $n=2000$, the observed narrow oscillations and marginal shifts in $R^2$ (such as the transition from $0.737$ to $0.808$ for \texttt{ER-5-0} at $d=11$) reflect standard asymptotic behavior rather than structural instability. In smaller samples ($n=200$) coupled with dense local configurations, ordinary least squares (OLS) estimations are susceptible to minor small-sample overfitting biases and random noise fluctuations. Expanding the horizon to $n=2000$ invokes the law of large numbers, allowing the empirical sample $R^2$ to cleanly converge toward its true population parameter dictated by the underlying coefficient matrix $W_{\text{true}}$ and error variance $\sigma^2$. This asymptotic alignment provides a more precise and stable empirical characterization of the robust causal linear pathways surviving the clique-free filtering process, particularly within low-dimensional saturated boundaries.

\section{Callback Events, Timing, and MIPSOL Interpretation}
\label{sec:callback}
\begin{figure*}[t]
\centering
\includegraphics[width=0.98\textwidth]{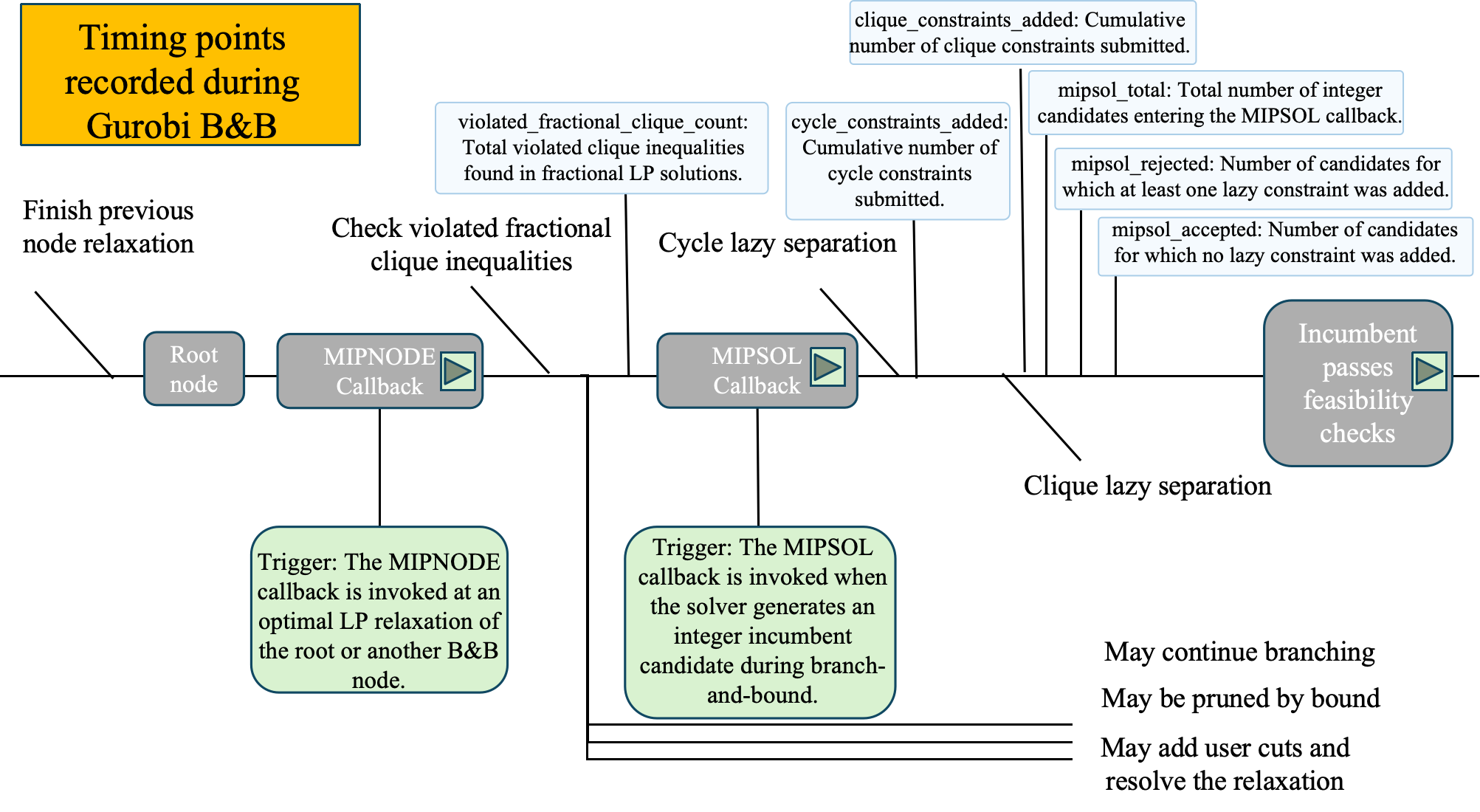}
\caption{\textbf{Timing points in branch-and-bound search.}
For each integer candidate, the callback first performs cycle and clique
separation and may submit lazy constraints. After control returns to Gurobi,
the solver continues branch-and-bound search until a subsequent candidate,
bound update, or termination event.}
\label{fig:time_point}
\end{figure*}

In this section, we formally define the \texttt{MIPSOL}, \texttt{MIPNODE}, cut, bound, timing, and checkpoint quantities used throughout our analysis, illustrating exactly where they are recorded during the branch-and-bound search in Figure~\ref{fig:time_point}. To support the empirical results presented in the main manuscript regarding the distinction between cycle-only, clique-only, and jointly violating candidates in the callback analysis, as well as the fractional user cuts findings.  

The callback mechanism operates via two main event types: \texttt{MIPSOL} and \texttt{MIPNODE}, corresponding to integer candidate solutions and node LP relaxations, respectively. As distinguished in Table~\ref{tab:mipsol-mipnode}, these events delineate the computational workload triggered inside a callback from the solver's subsequent independent search. At a \texttt{MIPSOL} event, the callback intercepts an integer candidate, verifies cycle and clique compliance, and rejects infeasible configurations by submitting lazy constraints via \texttt{cbLazy()}. Once the callback returns, Gurobi resumes its standard branch-and-bound routine until the next candidate is generated. Consequently, callback duration precisely measures the computational cost of separation and constraint submission, whereas the remaining runtime reflects baseline model processing, LP relaxations, branching decisions, primal heuristics, and node search.

When analyzing these events, callback frequencies, detected violations, and successfully submitted constraints must be treated as distinct metrics. A \texttt{MIPSOL} record represents a downstream outcome of the global search process rather than a direct measure of immediate relaxation strength. Similarly, the volume of \texttt{MIPNODE} invocations, the frequency of violated fractional clique inequalities, and the total of submitted user cuts describe separate structural dimensions of the optimization trajectory. 

\paragraph{Interpreting \texttt{MIPSOL} Counts.}
A larger \texttt{MIPSOL} count is neither inherently beneficial nor detrimental; its utility must be evaluated jointly with incumbent improvements, dual bound progress, rejection rates, and node pruning efficiency. An elevated count is productive when accepted candidates repeatedly update the incumbent, thereby tightening the objective cutoff and allowing more branch-and-bound nodes to be pruned. Conversely, it becomes an optimization bottleneck when lazy cycle or clique constraints continuously reject candidates without advancing either the primal or dual bounds. For example, under the configuration $M=10$ with seed~3, \texttt{Clique} generated 741 \texttt{MIPSOL} events compared to 735 for \texttt{ExDBN}. However, both models converged to nearly identical objective/bound pairs ($24.8/8.2$ versus $24.6/8.3$) and relative gaps ($0.67$ versus $0.66$). In this scenario, the six additional \texttt{MIPSOL} events yielded negligible optimization progress, primarily incurring redundant separation and search overhead.

Incorporating fractional user cuts introduces a severe performance bottleneck. The intensive separation overhead outweighs its dual-bound strengthening benefits, with the \texttt{Root + sparse non-root} configuration spending a substantial portion of runtime ($41.09\%$ on \texttt{ER-4-0} and $47.38\%$ on \texttt{ER-5-0}) confined within callbacks.
Under the fixed $1,000$-second budget, the computational burden of active separation drastically reduces node exploration speeds. For instance, the throughput drops from $2,110$ to $1,297$ nodes/second compared to the baseline \texttt{No cbCut} configuration, yielding negligible improvements or even slight degradation in the final relative optimality gap.
These trends demonstrate that exhaustive or heuristic fractional cut generation at the root or non-root nodes fails to alleviate the dual-bound bottleneck efficiently. Consequently, standard integer-candidate lazy constraints via \texttt{MIPSOL} remain a computationally superior strategy over fractional \texttt{MIPNODE} separation for this causal structure learning framework.

\begin{table}[t]
\centering
\small
\caption{\textbf{Operational distinction between \texttt{MIPSOL} and \texttt{MIPNODE} events.}}
\label{tab:mipsol-mipnode}
\begin{tabular}{p{0.18\columnwidth} p{0.36\columnwidth} p{0.28\columnwidth}}
\toprule
\textbf{Event Type} & \textbf{Description} & \textbf{Callback Action} \\
\midrule
\texttt{MIPSOL} &
Triggered when the solver finds an integer candidate, originating either from a branch-and-bound node or an internal primal heuristic. &
\texttt{cbLazy()} may reject the candidate by appending valid lazy constraints. \\
\midrule
\texttt{MIPNODE} &
Triggered when a continuous LP relaxation is available at a branch-and-bound node. The solution is typically fractional. &
\texttt{cbCut()} may submit globally valid user cuts; root-node diagnostics are also logged. \\
\bottomrule
\end{tabular}
\end{table}

\section{Additional Results}
\label{sec:additional_results}

This section presents the supplementary figures and detailed experimental protocols for Improvements A--G discussed in the main manuscript. 

\begin{figure*}[t]
\centering
\includegraphics[width=0.48\textwidth]{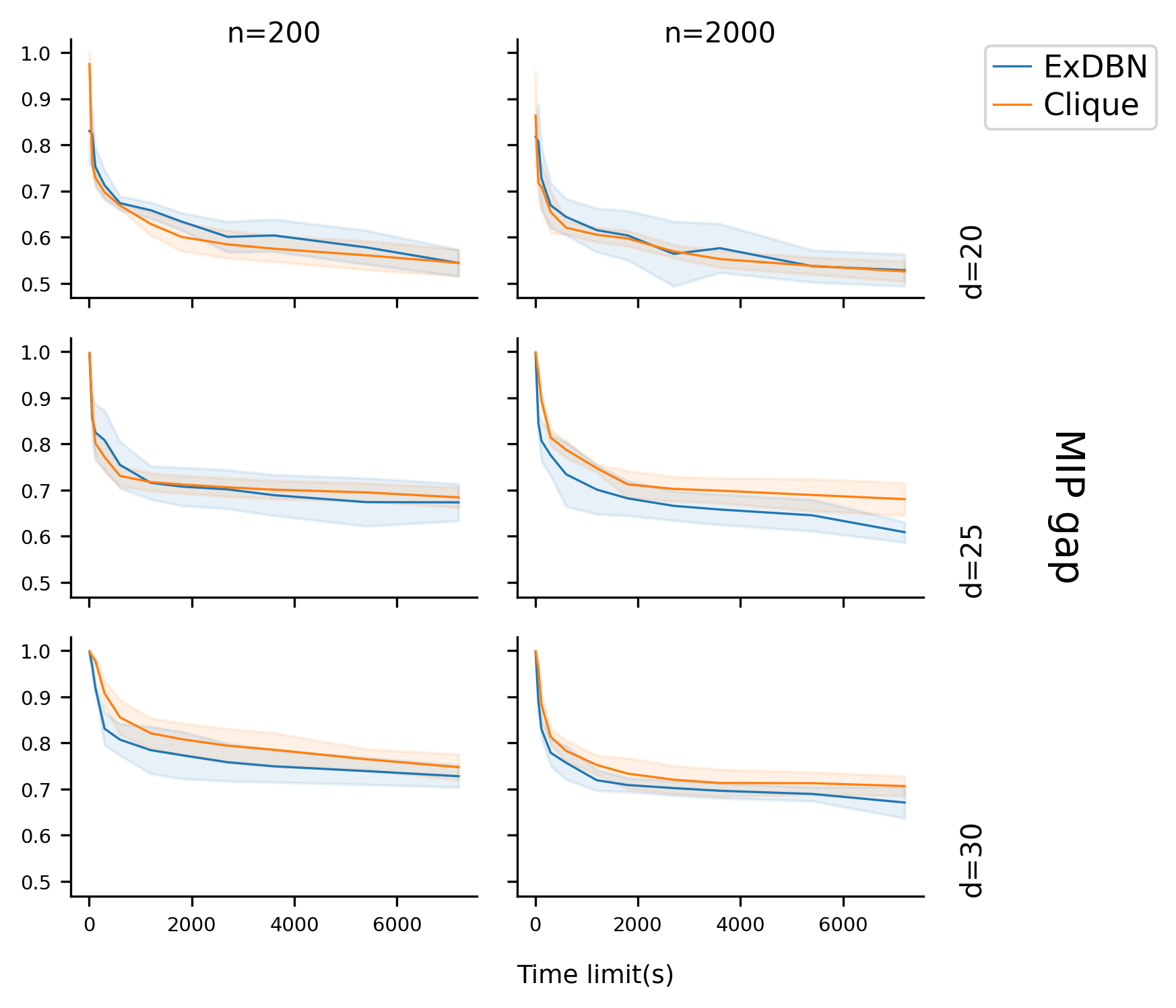}
\includegraphics[width=0.48\textwidth]{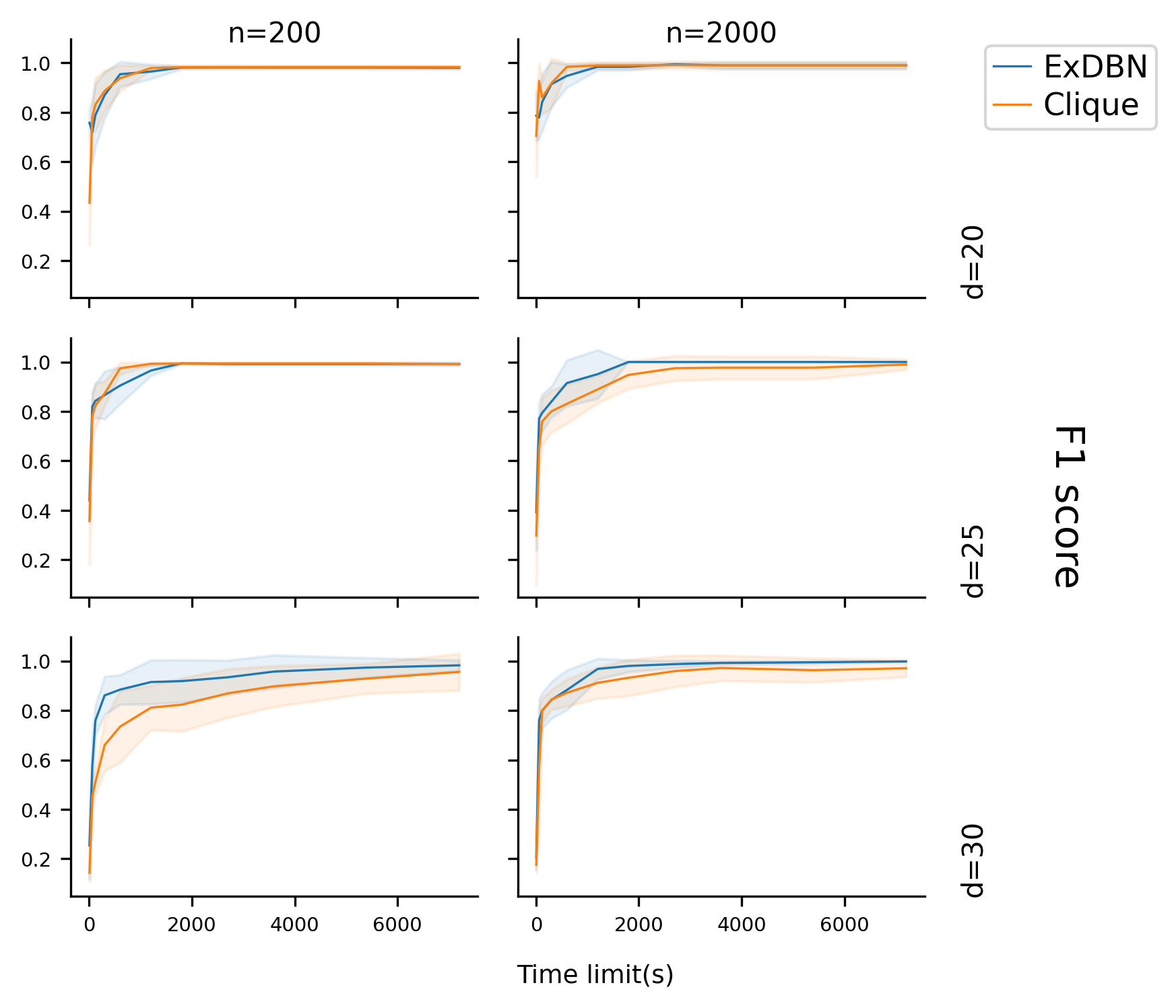}
\caption{\textbf{Relative MIP gap and F1 score for Clique and ExDBN across
time limits and dimensions $d$.} Lines show averages over seeds and shaded
regions report variability across runs.}
\label{fig:results_timelimit_gap_and_f1}
\end{figure*}

Figure~\ref{fig:results_timelimit_gap_and_f1} illustrates that the relative MIP gap often continues to decrease monotonically long after the F1 score has stabilized. This trajectory profile reflects a fundamental decoupling between bounding progress and structural recovery: the gap evolves whenever either the primal or dual bound improves, whereas the F1 score adapts exclusively when a newly accepted incumbent alters the underlying graph topology. Consequently, these trends demonstrate that causal graph recovery quality is bounded by computational convergence, yet further numerical optimization faces diminishing returns regarding structural refinement once the search layout settles. 

\begin{table*}[t]
\centering
\small 
\renewcommand{\arraystretch}{1.2} 
\setlength{\tabcolsep}{8pt} 

\begin{tabularx}{\textwidth}{@{} p{3.5cm} p{4.0cm} X @{}}
\toprule
\textbf{Improvements} &
\textbf{Methods} &
\textbf{Experiments} \\
\midrule

\multirow[c]{2}{3.5cm}{
\raggedright\textbf{Reduce redundant searches}
}
& \textbf{A. Delayed Separation} & 
Reduce clique separation, increase solving nodes per sec and close gap. \\
\cmidrule(lr){2-3}

& \textbf{C. Repair Incumbent} & 
Repair an infeasible MIPSOL, locally refitting support with ridge/QR. \\
\midrule

\multirow[c]{2}{3.5cm}{
\raggedright\textbf{Control adding constraint number}
}
& \textbf{B. Controlled Cut Addition} & 
First oversized cliques (original), First size-min-k+1-clique, most-violated size-min-k+1-clique. \\
\cmidrule(lr){2-3}

& \textbf{F. Stronger Clique Cuts} & 
Top-3-size-min-k+1-cliques, Top-5-size-min-k+1-cliques, All-unique-size-min-k+1-cliques (adding $\binom{8}{6}$ constraints). \\
\midrule

\multirow[c]{2}{3.5cm}{
\raggedright\textbf{Strengthen the LP relaxation and dual bound}
}
& \textbf{D. Fractional Separation} & 
Inspect the fractional MIPNODE relaxation and submit valid violated clique inequalities through \texttt{cbCut()}. \\
\cmidrule(lr){2-3}

& \textbf{E. Big-\(M\) Relaxation} & 
\(M\in\{10,15,20,25,50,100\}\) and edge-specific \(M_{ij}\) linking bounds. \\
\midrule

\raggedright\textbf{Checkpoints}
& \textbf{G. Time-Limit Optimization} & 
Run one continuous solve and record metrics at fixed checkpoints. \\
\bottomrule
\end{tabularx}

\caption{\textbf{Improvements and experimental directions.} 
Variants \texttt{B1}--\texttt{F2}, \texttt{B3}, \texttt{C3}, and \texttt{E2} are systematically compared in the main manuscript.}
\label{tab:improvement-directions}
\end{table*}

\paragraph{Stability Diagnostics}
\label{sec:stability_diagnostics}

In addition to the clique-constrained MILP formulation, we evaluate an edge-limit MILP variant to analyze stability, where structural complexity is controlled by static linear constraints imposed at model construction time rather than dynamically separated clique cuts. These constraints restrict both the in-degree and out-degree of every node, remaining invariant throughout the optimization trajectory. To match our experimental envelope where the maximum dimensionality reaches $d=35$, we evaluate degree bounds ranging from $0$ to $35$ formalized as: 
\begin{equation}
\sum_{i \in V} z_{ij} \le d_{\text{in}}^{\max}, \quad \forall j \in V, \label{eq:in_degree}
\end{equation}
\begin{equation}
\sum_{j \in V} z_{ij} \le d_{\text{out}}^{\max}, \quad \forall i \in V, \label{eq:out_degree}
\end{equation}
where $z_{ij} \in \{0,1\}$ denotes the binary edge-selection variable associated with the learned adjacency matrix. By enforcing a global degree envelope through a priori linear structures rather than via runtime lazy cuts, this formulation establishes a controlled environment to study algorithmic behavior, where relaxing the edge boundaries expands the feasible search space of graph topologies but simultaneously exacerbates the combinatorial complexity of the underlying branch-and-bound tree.

\begin{figure*}[t]
\centering

\begin{subfigure}{0.48\textwidth}
    \centering
    \includegraphics[width=0.98\textwidth]{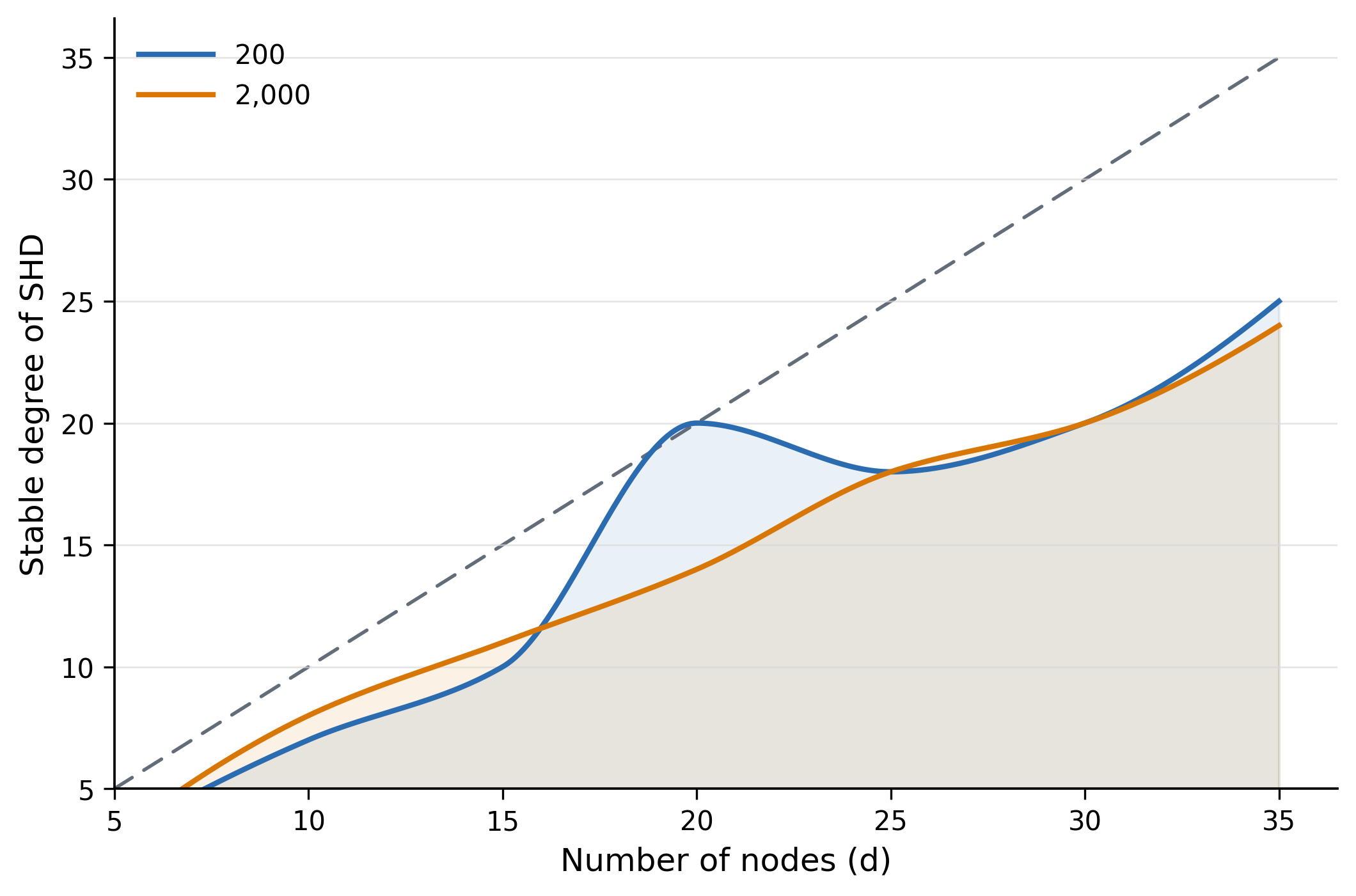}
    \caption{Structural Hamming Distance (SHD) trends}
    \label{fig:stability_shd_trends}
\end{subfigure}
\hfill
\begin{subfigure}{0.48\textwidth}
    \centering
    \includegraphics[width=0.98\textwidth]{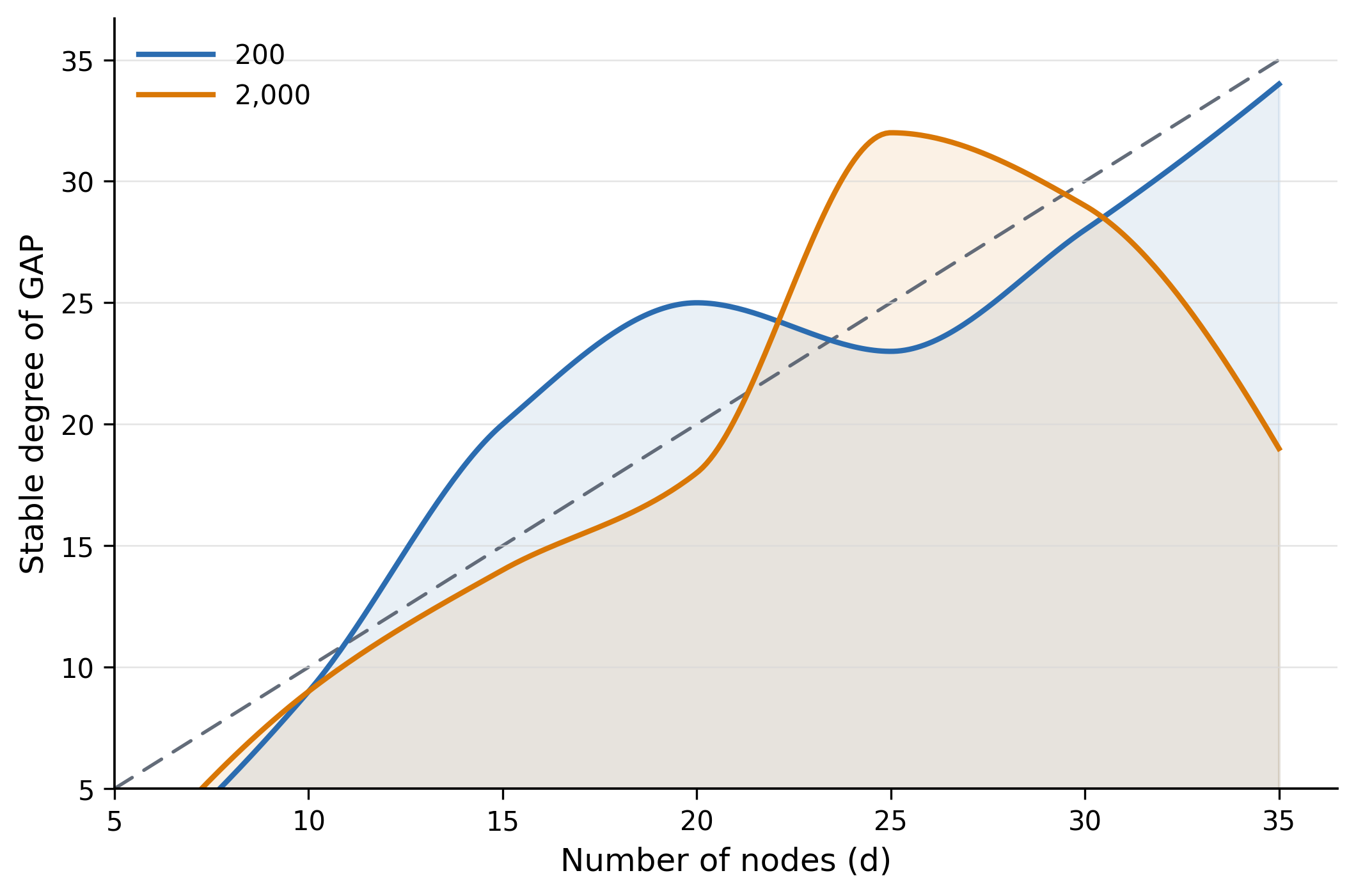}
    \caption{Relative MIP gap progression}
    \label{fig:stability_gap_trends}
\end{subfigure}

\vspace{1.2em} %
\begin{subtable}{0.90\textwidth}
\centering
\small
\renewcommand{\arraystretch}{1.1} %
\setlength{\tabcolsep}{10pt}      %

\resizebox{\linewidth}{!}{
\begin{tabular}{l cccc cccc}
\toprule
\multirow{3}{*}{\textbf{Nodes ($d$)}} & \multicolumn{4}{c}{\textbf{Minimal Stable SHD Bound}} & \multicolumn{4}{c}{\textbf{Minimal Stable GAP Bound}} \\
\cmidrule(lr){2-5} \cmidrule(lr){6-9}
& \multicolumn{2}{c}{\textbf{Sample $n=200$}} & \multicolumn{2}{c}{\textbf{Sample $n=2000$}} & \multicolumn{2}{c}{\textbf{Sample $n=200$}} & \multicolumn{2}{c}{\textbf{Sample $n=2000$}} \\
\cmidrule(lr){2-3} \cmidrule(lr){4-5} \cmidrule(lr){6-7} \cmidrule(lr){8-9}
& \textbf{Degree} & \textbf{SHD Value} & \textbf{Degree} & \textbf{SHD Value} & \textbf{Degree} & \textbf{GAP Value} & \textbf{Degree} & \textbf{GAP Value} \\
\midrule
5  & 3  & 3.0  & 3  & 2.0  & 1  & 0.0000 & 1  & 0.0000 \\
10 & 7  & 7.0  & 8  & 5.0  & 9  & 0.0100 & 9  & 0.0099 \\
15 & 10 & 11.5 & 11 & 2.0  & 20 & $0.27^*$ & 14 & $0.29^*$ \\
20 & 20 & 2.0  & 14 & 0.0  & 25 & $0.61^*$ & 18 & $0.56^*$ \\
25 & 18 & 1.0  & 18 & 0.0  & 23 & $0.63^*$ & 24 & $0.65^*$ \\
30 & 20 & 0.0  & 20 & 0.0  & 28 & $0.79^*$ & 29 & $0.78^*$ \\
35 & 25 & 1.0  & 24 & 0.0  & 34 & 0.8517 & 19 & $0.82^*$ \\
\bottomrule
\end{tabular}
}
\caption{Empirical thresholds for structural and optimization stability}
\label{tab:stability_thresholds_clean}
\end{subtable}

\vspace{0.6em}

\caption{\textbf{Edge-limit stability trajectories and empirical plateaus across dimension regimes.} Panels (a) and (b) trace the global continuous optimization dynamics, while the unified data subtable (c) catalogs the exact minimal degree boundaries ($d_{\max}$) beyond which additional metric relaxation yields marginal structural variation. Asterisks ($*$) denote approximate plateaus resolved to two decimal places. These thresholds are descriptive diagnostic references and do not represent formal asymptotic convergence guarantees.}
\label{fig:stability_diagnostics_unified}
\end{figure*}

The continuous empirical trajectories and exact structural plateaus resulting from this setup are collectively systematized in Figure~\ref{fig:stability_diagnostics_unified}. As illustrated in the graphical profiles (a and b), relaxing the edge limits initially enhances statistical recovery if a tighter bound previously excluded true causal relationships, yet it eventually slows down dual bound progression and optimality-gap closure, capturing a clear statistical--computational trade-off. The empirical stability boundaries cataloged in the unified subtable (c) further confirm this trade-off; for lower dimensions ($d \le 15$), expanding the horizon from $n=200$ to $n=2000$ cleanly reduces the minimal stable SHD bound from $11.5$ to an exceptionally tight value of $2.0$. 
These combined diagnostics demonstrate that causal graph recovery quality is fundamentally driven by computational convergence, showing that further structural refinement faces diminishing returns long before the full optimality gap is closed.

\section{Notation Table}
\label{sec:notation}
Notations and terminologies are listed in Table~\ref{tab:Notation_Table}.
\begin{table*}[t]
\centering
\small
\renewcommand{\arraystretch}{1.01}
\setlength{\tabcolsep}{5pt}
\begin{tabular}{p{3.5cm}p{2.5cm}p{9.2cm}}
\toprule
\textbf{Concept} & \textbf{Notation / Gurobi term} & \textbf{Definition} \\

\midrule
Branch-and-bound & B\&B & Tree-search for mixed-integer optimization; nodes define subproblems whose continuous relaxations supply bounds. \\
Search tree & $\mathcal{T}$ & Collection of generated B\&B nodes; a root-to-node path records the branching decisions. \\
Node queue & $\mathcal{Q}$; \texttt{MIP\_NODLFT} & Unexplored B\&B nodes remaining in the solver queue. \\
Current node & $v\in\mathcal{T}$ & The B\&B subproblem currently processed. Its LP relaxation can be fractional. \\
Callback & \texttt{MIPSOL}; \texttt{MIPNODE} & User code at solver events. \texttt{MIPSOL} inspects integer candidates; \texttt{MIPNODE} inspects fractional relaxations. \\
Incumbent & $z^{\mathrm{inc}}$; \texttt{MIP\_OBJBST} & Best globally feasible integer solution found so far. \\
Primal bound / upper bound & $UB=f(z^{\mathrm{inc}})$ & Objective value of the current incumbent (minimization convention). \\
Dual bound / lower bound & $LB$; \texttt{MIP\_OBJBND} & Global lower bound on the optimal objective, derived from active node relaxations. \\
Relative optimality gap & $\displaystyle\frac{UB-LB}{\max\{|UB|,\epsilon\}}$ & Normalized separation between primal and dual bounds. %
\\
Clique separation & $\mathcal{S}_{\mathrm{clique}}$ & Detection of violated cliques. \texttt{cbLazy()} rejects candidates at \texttt{MIPSOL}; \texttt{cbCut()} adds cuts at \texttt{MIPNODE}. \\
Clique lazy constraint & \texttt{cbLazy()} & Dynamically generated constraint enforcing integer feasibility without directly strengthening fractional relaxations. \\
Clique user cut & \texttt{cbCut()} & Globally valid inequality submitted at fractional MIPNODE relaxations to strengthen them. \\

\midrule
Samples, variables, and lags & $n$, $d$, $p$ & Numbers of observations, graph variables, and temporal lags ($p=0$ represents static data). \\
Data matrices & $X$; $Y^{(t)}$ & Contemporaneous and lag-$t$ data matrices. \\
Coefficient matrices & $W$; $A^{(t)}$ & Contemporaneous adjacency matrix ($W_{ij}\neq 0 \implies i\rightarrow j$) and lag-$t$ coefficient matrix. \\
Edge-selection variable & $z_{ij}\in\{0,1\}$ & Binary indicator whether directed edge $i\rightarrow j$ is selected. \\
Residual and absolute-loss & $r_{ij}$; $u_{ij}$ & Regression residual and its linearized absolute-value auxiliary variable. \\
Regularization & $\lambda_1,\lambda_2$; $\mathrm{Reg}_W,\mathrm{Reg}_A$ & Penalty weights and regularizers for contemporaneous and lagged coefficients. \\
Big-\(M\) bound & $M$ & Bound in $-Mz_{ij}\leq W_{ij}\leq Mz_{ij}$. Tighter valid values strengthen relaxation
. \\

\midrule
Graph& $G=(V,E)$ & Selected directed graph.\\
Support &  & Edge skeleton. \\
Cycle & $C$ & Directed cycle
. \\
Clique and clique bound & $Q$; $k$; $\omega(G)$ & Clique $Q$, maximum permitted clique size $k$, and graph clique number ($\omega(G)\leq k$). \\
Clique-pair set & $P(Q)$ & Unordered pairs in $Q$, with $|P(Q)|=\binom{|Q|}{2}$. \\
Minimal forbidden subset & $S\subseteq Q$, $|S|=k+1$ & Smallest subset violating the clique bound. \\
Clique-cut violation & $v(S)$ & Violation score used by B3 to rank minimal $(k+1)$ clique inequalities. \\
B3 & Most-violated minimal clique cut & MIPSOL policy submitting highest-ranked minimal $(k+1)$ inequalities via \texttt{cbLazy()} under cut limit. \\
C3 & QR-based incumbent repair & Primal heuristic that deletes violating edges, refits columns via QR, and submits via \texttt{cbSetSolution()}. \\
Recovery metrics & SHD; F1 & SHD measures graph-edit distance to ground truth; F1 is the harmonic mean of precision and recall. \\
\bottomrule
\end{tabular}

\caption{\textbf{Notation used throughout the paper.} The table consolidates
optimization terminology, statistical-model notation, and graph, method, and
evaluation notation.}
\label{tab:Notation_Table}
\end{table*}

\newpage

\end{document}